\documentclass[letterpaper, 10 pt, conference]{ieeeconf}  % Comment this line out if you need a4paper
\pdfoutput=1

\IEEEoverridecommandlockouts                              % This command is only needed if 
\usepackage{booktabs}

\usepackage{tabu}
\usepackage{cite}
\usepackage{multirow}
\usepackage{flushend}

\usepackage[english]{babel}
\usepackage{lipsum}

\usepackage{amsmath}
\usepackage{amssymb}

\usepackage[ruled,vlined]{algorithm2e}
\usepackage{graphics} % for pdf, bitmapped graphics files
\usepackage{graphicx,color}

\usepackage{color, colortbl, soul}
\usepackage{xcolor}

\usepackage{todonotes}

\usepackage{amsfonts}
\usepackage{siunitx}

\usepackage{pifont}
\newcommand*\colourcheck[1]{%
  \expandafter\newcommand\csname #1check\endcsname{\textcolor{#1}{\ding{52}}}%
}

\newcommand*\colourcross[1]{%
  \expandafter\newcommand\csname #1cross\endcsname{\textcolor{#1}{\ding{55}}}%
}

\colourcross{red}

\colourcheck{blue}
\colourcheck{green}
\colourcheck{red}

\title{\LARGE \bf
Data-efficient learning\\ of object-centric grasp preferences
}

\author{Yoann Fleytoux$^{1}$, Anji Ma$^{1,2}$, Serena Ivaldi$^{1}$, Jean-Baptiste Mouret$^{1}$%
\thanks{*This work was supported by the CHIST-ERA grant HEAP (CHIST-ERA-17-ORMR-003).}% <-this % stops a space
\thanks{$^{1}$Universit\'e de Lorraine, CNRS, Inria, F-54600, France}
\thanks{$^{2}$School of Mechatronical Engineering, Beijing Institute of Technology, Beijing 100081, China}
\thanks{\tt\small jean-baptiste.mouret@inria.fr}
 }

\begin{document}
\thispagestyle{empty}
\pagestyle{empty}

\maketitle

%===============================================================================

\begin{abstract}
Grasping made impressive progress during the last few years thanks to deep learning. However, there are many objects for which it is not possible to choose a grasp by only looking at an RGB-D image, might it be for physical reasons (e.g., a hammer with uneven mass distribution) or task constraints (e.g., food that should not be spoiled). In such situations, the preferences of experts need to be taken into account.

In this paper, we introduce a data-efficient grasping pipeline (Latent Space GP Selector --- LGPS) that learns grasp preferences with only a few labels per object (typically 1 to 4) and generalizes to new views of this object. Our pipeline is based on learning a latent space of grasps with a dataset generated with any state-of-the-art grasp generator (e.g., Dex-Net). This latent space is then used as a low-dimensional input for a Gaussian process classifier that selects the preferred grasp \emph{among those proposed by the generator}.

The results show that our method outperforms both GR-ConvNet and GG-CNN (two state-of-the-art methods that are also based on labeled grasps) on the Cornell dataset, especially when only a few labels are used: only 80 labels are enough to correctly choose 80\% of the grasps (885 scenes, 244 objects). Results are similar on our dataset (91 scenes, 28 objects).

\end{abstract}

% \keywords{Grasping, Gaussian processes} 

%===============================================================================

%%%%%%%%%%%%%%%%%%%%%%%%%%%%%%%%%%%%%%%%%%%%%%%%%%%%%%%%%%%%%%%%%%%%%%%%%%%%%%%%
\section{INTRODUCTION}

% \begin{wrapfigure}[21]{r}{0.35\linewidth}
%   \centering
%   \vspace{-10pt}
%   \includegraphics[width=\linewidth]{./figure/hammer.pdf}
%   \caption{Grasp generated by Dex-Net 4.0 \cite{Mahlereaau4984} for a toy hammer and with our method with only 2 labels (one positive label, in green, and one negative label, in red). See section \ref{sec:qualitative}.}
%   \label{fig:hammer}
% \end{wrapfigure}

Robots are often tasked with grasping objects from a box or a conveying belt, especially in industrial settings \cite{prattichizzo2016grasping}. Thanks to the recent advances in deep learning, they can now grasp unknown objects with success rates that exceed 90\%~\cite{du2021vision}. To do so, they learn the relationship between the shape of the objects and the most appropriate grasp.

Nevertheless, the best grasp according to the shape is not always the grasp that should be favored. For instance, a hammer should usually be taken by the head because the mass distribution is not uniform, whereas the shape suggests that it should be taken by the handle. Similarly, a sharp knife should usually be grasped by the handle to avoid damaging the gripper; but it might be necessary to grasp it by the blade if the robot needs to give it to an operator. For some objects, it might be legally forbidden to grasp them in some way, for example in the food industry; in some other cases, the objects might break or be scratched when grasped in the wrong way, for example, by grasping prescription glasses by a lens, or the objects require taking specific properties into account, like fresh paint on a sub-part. Importantly, the grasp that should be preferred does not always help grasping; instead, it is often a constraint of the task.

Hence, expert knowledge for grasping specific objects is required in many situations; but labeling thousands of images is time-consuming and would need to be performed for each application (e.g., for a hammer factory, then for a knife warehouse, etc.). Our grasp pipeline aims at learning object-centric grasping preferences with very few examples (typically 1 to 4 per object) and generalizing to other views of the same object. In this work, we target single objects on a simple background, for instance, objects on a conveyor belt in industrial settings.

Our main insight is that we can leverage generic grasp generators to make learning preferences data-efficient in two ways: (1) they can be run on a large dataset of images to learn a latent space for grasps, and (2) they can be used to generate grasp candidates so that a preference-based classifier only has to choose among good grasps. In other words, a grasp generator takes the needles out of the haystack, and a classifier only has to choose the best needle in a low-dimensional space.

\begin{figure}[t]
   \centering
   \includegraphics[width=\linewidth]{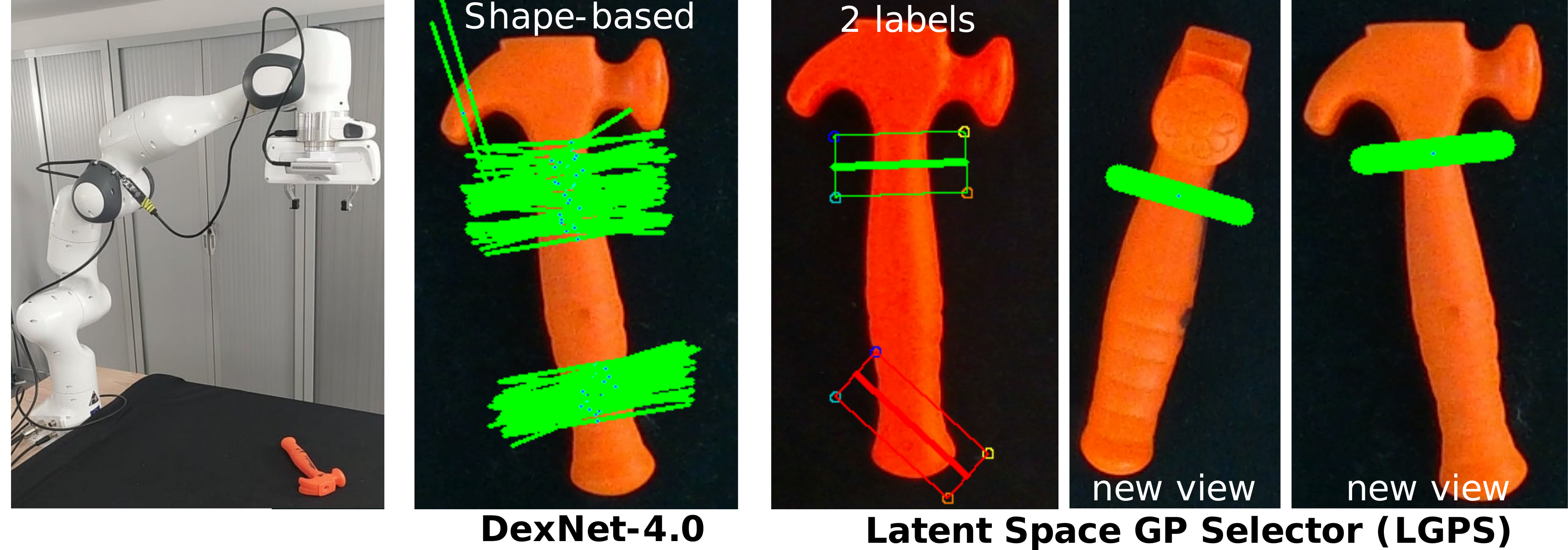}
   \caption{Grasp generated by Dex-Net 4.0 \cite{Mahlereaau4984} for a toy hammer and with our method with only 2 labels (one positive label, in green, and one negative label, in red). See section \ref{sec:qualitative}.}
   \label{fig:hammer}
\end{figure}

Our main assumption is that we have access to a set of passively-obtained RGB-D images on which a shape-based grasp generator can be run; we consider this dataset as ``large and cheap''. We represent the generated grasps using rotated and adjusted image patches that embed the grasp and its context in a single input \cite{DBLP:conf/rss/MahlerLNLDLOG17}, which makes it possible to learn a low-dimensional representation of both the grasps and their context with a Variational Auto-Encoder (VAE) \cite{vae_kingma_2014_iclr,DBLP:journals/corr/abs-1804-03599}. We then use this low-dimensional representation to train/query a Gaussian Process (GP) classifier \cite{DBLP:conf/aistats/HensmanMG15,DBLP:books/lib/RasmussenW06} that \emph{filters} the grasps generated by a grasp generator, which are often already effective, according to the preferences of the expert. Importantly, the expert can give both ``positive labels'', that is, grasps that should be favored (e.g., ``this is how this should be done''), and ``negative labels'', that is, grasps that should be avoided (e.g., ``do not do this''). We call our method ``Latent Space GP Selector'' (LGPS).

Once trained, our pipeline generates grasp candidates, encodes them in the latent space, then queries a Gaussian process classifier to know the preference of the expert; the robot executes this grasp with a standard planning algorithm \cite{quigley2009ros}. For the preference training, the grasp selected by the expert is encoded to the same latent space and the Gaussian process classifier is updated.

%\yoann{pmlr-v108-fortuin20a  utilizes Variational Autoencoders with Gaussian Process prior for time series imputation.}
While VAE and GP have been combined together in different fields (e.g.,~\cite{yoo2017variational} for videos), our main contribution is the combination of grasp generators (which can, for instance, be based on deep learning) with an image-based grasp representation to learn a latent space of grasps in an unsupervised way. We show that our pipeline makes it possible to learn grasps that are about $80\%$ consistent with the expert labels with less than one example per object on the Cornell dataset \cite{DBLP:conf/icra/JiangMS11} (885 scenes, 244 objects) and on our own dataset (91 scenes, 28 objects).

\section{Related work}

% \todo[inline]{The aim of a related work section is to situate your contribution with respect to other people’s work.  You want to give the reader an overview of papers that are solving similar problems or using similar methods.  You also want to tell the reader why your contribution is new, different, and better.  A good startegy to learn how to write a related work section is described here: \url{https://h2r.cs.brown.edu/writing-a-technical-paper/}. I find useful to organize the related works into groups of papers by theme, method or application. You need to make sure you know the papers that are most relevant for your problem, and situate these important papers in the galaxy of the papers related to the big field. If the field is too big, for example ``grasping'' is a very big field, try to come up with a synthetic view of the main approaches and situate where your work is. Do not underestimate the time for writing this section: it can be several weeks, especially if you don't have a bibtex file with all the references of the papers.}

% \todo[inline]{Identify the gap that your paper is filling.}

% \yoann{review https://arxiv.org/pdf/1905.06658.pdf}

Vision-based robotic grasp planning methods can be classified into analytical \cite{DBLP:conf/icra/FerrariC92} and data-driven approaches \cite{DatadrivenGrasp2013Kragic}. Analytical approaches require information about the physical properties of the manipulated objects (shape, mass, center of mass, friction coefficient), which is possible only on well-controlled manufacturing scenarios. By contrast, data-driven approaches attempt to generalize to unknown objects by being trained on datasets of grasp examples.

Data-driven approaches mainly differ in the kind of dataset they use. A recently successful idea is to create large synthetic databases from 3D models and simulations \cite{DBLP:conf/iros/JohnsLD16,DBLP:conf/corl/GualtieriP18, DBLP:conf/rss/MahlerLNLDLOG17,DBLP:conf/iros/ZengSWLRF18,DBLP:conf/iccv/MousavianEF19,DBLP:conf/corl/ViereckPSP17}. The strength of these datasets is their size (millions of objects), but they currently only take into account the 3D shape (via depth data): they  assume a uniform mass distribution of objects. As they strongly rely on simulation, they often need adaptation methods to be effective with real robots \cite{DBLP:conf/icra/BousmalisIWBKKD18}; on the other hand, automatically gathering sufficient training data through trial and error with real robots \cite{DBLP:conf/icra/PintoG16, DBLP:journals/ijrr/LevinePKIQ18} is highly time-consuming and does not necessarily provide better results (e.g., 50,000 trials and 700 hours of robot use in  \cite{DBLP:conf/icra/PintoG16, DBLP:journals/ijrr/LevinePKIQ18}). Human feedback was used \cite{DBLP:journals/arobots/DanielKVM015, DBLP:conf/icra/PinslerAO0N18} such that the robot only performs hundreds of grasps during task learning, but these manipulations can cause safety concerns when dealing with particular objects where certain grasps must be avoided.

The second kind of data-driven approach learns from a dataset in which objects have been labeled by experts to specify where/how to grasp objects.
Several studies have been proposed to exploit data from captured expert manipulations such as \cite{DBLP:journals/scirobotics/MigliozziLFS19}, which learns interactions from videos of experts, or \cite{DBLP:conf/hri/PraveenaSMG19, DBLP:journals/corr/abs-1912-04344}, which uses custom handheld devices to collect grasping demonstration. Nevertheless, many recent studies \cite{chu2018deep, DBLP:conf/rss/MorrisonLC18, DBLP:journals/corr/abs-1909-04810} use the Cornell dataset \cite{DBLP:conf/icra/JiangMS11} or similarly acquired datasets \cite{DBLP:conf/icra/ZengSYDHBMTLRFA18,de2020learning} which provide thousands of grasp locations labeled by humans.

These grasp demonstrations have been used to infer an evaluation function that ranks grasp candidates according to the expert specifications \cite{lenz2015deep,DBLP:conf/icra/PintoG16, park2018classification}. Recently Generative Grasping CNN (GG-CNN) \cite{DBLP:conf/rss/MorrisonLC18}, Generative Residual Convolutional Neural Network (GR-ConvNet) \cite{DBLP:journals/corr/abs-1909-04810} and other work use hand-designed labels to generate pixel-wise grasp affordance map. The grasping task is then similar to semantic segmentation, in which some parts of the object are deemed graspable whereas some others have to be avoided. 

While generic grasping of unknown objects needs to use generic datasets, labeling specific datasets for specific objects is highly time-consuming. A few papers therefore focus on learning grasp preferences with as few ``demonstrations'' as possible \cite{van2018learning}. In particular, \cite{DBLP:conf/ro-man/HelenonBNTG20} proposes a CNN, pixel-wise segmentation pipeline that predicts authorized grasping locations from depth image while avoiding prohibited locations. The demonstrations are gathered by applying colored tape to the operator's fingers, who thus can demonstrate parallel grasps. 
This pipeline was evaluated with 5 types of industrial objects (socket wrenches, pliers, light bulbs, cups, and screws), using between 1 and 3 demonstrations for training. To allow CNN to be trained with so few labels, the authors apply heavy data augmentation and train on each type of object separately, and that allows them to reach between 70\% and 90\% success depending on the object. However, they report that their pipeline can at best grasp 2 objects with 2 different grasping strategies with the same neural network.

\begin{figure*}
   \centering
   \includegraphics[width=0.7\textwidth]{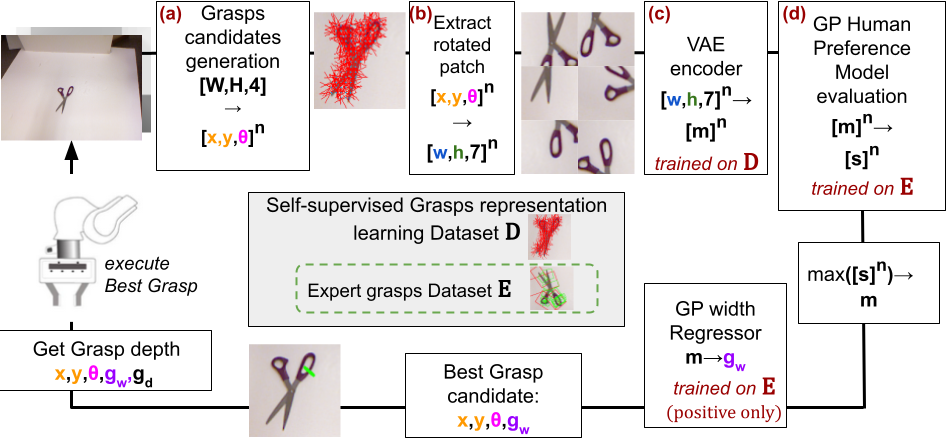}
   
   \caption{The LGPS grasping pipeline, assuming that the VAE has been trained before (Sec.~\ref{sec:vae}). From an RGB-D image, a grasp generator (Sec.~\ref{sec:grasp_generation}) creates grasp candidates as segments. These grasp candidates are represented as rotated patches centered on the middle of the segment (Sec.~\ref{sec:grasp_representation}). Each of them is fed to a VAE to get their latent representation, which is, in turn, the input of the GP classifier (Sec.~\ref{sec:GP}) to obtain the estimated probability of being selected by the expert; the grasp with the highest probability is selected. A second GP is queried to get the width of the gripper for the selected grasp. The ``depth'' of the grasp (the z-position of the gripper) is computed using the depth image.}
   \label{figure:Grasp_selection_process}

\end{figure*}

\section{Problem Statement}\label{sec:proble_statement}
Our main assumptions are: (1) the robot has an RGB-D camera; (2) all the objects can be grasped from the top; (3) we have access to a large unlabeled dataset of RGB-D images ($\mathbf{D}$); (4) we have access to a small labeled dataset ($\leq$ 4 per object) of grasps as either positive (good grasp) or negative (grasp to avoid) (dataset $\mathbf{E}$).

Our main objective is to learn to reproduce the grasps that the expert labeled as good for new views of objects that have been seen previously. Our secondary objective is to generalize to objects that have never been seen but that are close to those already labeled.

We evaluate the performance using the \emph{rectangle metric}~\cite{DBLP:conf/icra/JiangMS11}, which makes it possible to compare to previous works using published datasets \cite{DBLP:conf/icra/JiangMS11, DBLP:conf/iros/DepierreD018}. For this metric, grasps are represented as rectangles centered on the gripper's center position $(x,y)$, rotated according to the orientation $\theta$ $ \in \left[-\frac{\pi}{2}, +\frac{\pi}{2}\right]$, with a width that is equaling the gripper opening $l$, and a height that encodes the tolerance. Two grasps are compared by looking at how much the two corresponding rectangles overlap. More precisely, given a proposed grasp $GC$ and a ground truth grasp $GT$, the Intersection over Union (IoU, also called the Jaccard index) corresponds to the normalized area of intersection:

\begin{equation}\label{equation:IoU}
    IoU(GT,GC)  = \dfrac{| GT \cap GC|}{| GT \cup GC |}
\end{equation}

The rectangle metric $RM$ is equal to 1 if this overlapping area is above $0.25$ and the angular difference is below $\frac{\pi}{6}$:

\begin{equation}\label{equation:rectangle_metric}
    RM
    (GT,GC)= 
\begin{cases}
    1,&\text{if }IoU(GT,GC) > 0.25\\& \text{ and } \big| GT.\theta-GC.\theta\big| \leq \frac{\pi}{6}\\
    0,&\text{otherwise}
\end{cases}
\end{equation}

We compare our results to GR-Convnet \cite{DBLP:journals/corr/abs-1909-04810} and GC-CNN \cite{DBLP:conf/rss/MorrisonLC18} with the rectangle metric on the Cornell Dataset \cite{DBLP:conf/icra/JiangMS11} and our own dataset, which is focused on meaningful object-specific labels (scissors, hammer, etc.). Our method does not learn a height (tolerance) for the rectangles, therefore we have it set to $38$ pixels.

While the process described in the present paper is offline --- learning is performed with a labeled dataset (or a subset of this dataset) ---, we envision future \emph{online} deployments in which an expert corrects the robot online only when it is wrong.

\section{Method}\label{sec:method}

Our full algorithmic pipeline is illustrated in (Fig. \ref{figure:Grasp_selection_process}). The following subsections provide details about the respective components (a-d) of the pipeline.

% \begin{figure}[h]
%   \centering
%   \includegraphics[width=\textwidth]{./figure/pipeline.png}
   
%   \caption{The LGPS grasping pipeline, assuming that the VAE has been trained before (Sec.~\ref{sec:vae}). From an RGB-D image, a grasp generator (Sec.~\ref{sec:grasp_generation}) creates grasp candidates as segments. These grasp candidates are represented as rotated patches centered on the middle of the segment (Sec.~\ref{sec:grasp_representation}). Each of them is fed to a VAE to get their latent representation, which is, in turn, the input of the GP classifier (Sec.~\ref{sec:GP}) to obtain the estimated probability of being selected by the expert; the grasp with the highest probability is selected. A second GP is queried to get the width of the gripper for the selected grasp. The ``depth'' of the grasp (the z-position of the gripper) is computed using the depth image.}
%   \label{figure:Grasp_selection_process}
% \end{figure}

\subsection{Grasp candidate generation (a)}\label{sec:grasp_generation}
Given an RGB-D image, grasp candidates can be generated using computer vision techniques \cite{DBLP:journals/corr/abs-2001-02076}, or by randomly sampling them, or by using methods based on deep learning like Dex-Net \cite{DBLP:conf/icra/MahlerPHRLAKKKG16}, Generative Grasping CNN (GG-CNN) \cite{DBLP:conf/rss/MorrisonLC18} or Generative Residual Convolutional Neural Network (GR-ConvNet) \cite{DBLP:journals/corr/abs-1909-04810}. Here we use a grasp generator based on computer vision, which has proved to be both fast and effective in our experiments, once combined with a Gaussian process classifier (Sec.~\ref{sec:GP}), although a deep-learning generator could take its place easily. 

\begin{figure}[h]
   \centering
   \includegraphics[width=\linewidth]{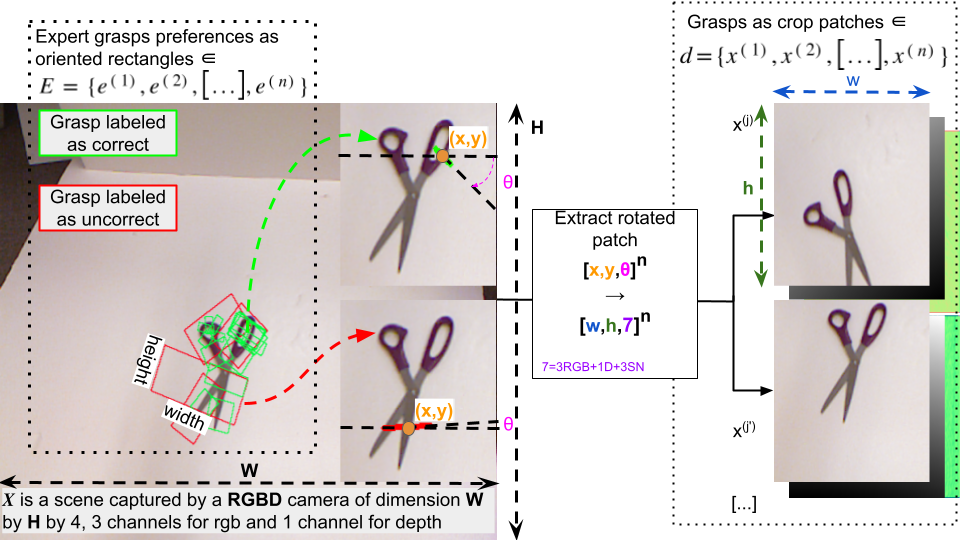}
  \caption{Grasps are represented as a rotated and cropped patch of the image that is centered on the grasp. This representation makes it possible to feed both grasps and their context to convolutional neural networks.}
  \label{figure:grasp_representation}
\end{figure}

% \begin{wrapfigure}[17]{hr}{9cm}
% \vspace{-25pt}
%   \centering
%   \includegraphics[width=9cm]{./figure/extract_patch_v2.png}
%   \caption{Grasps are represented as a rotated and cropped patch of the image that is centered on the grasp. This representation makes it possible to feed both grasps and their context to convolutional neural networks.}
%   \label{figure:grasp_representation}
% \end{wrapfigure}

%  We first extract the object with the GrabCut algorithm  \cite{DBLP:journals/tog/RotherKB04} (Suppl. Fig. \ref{figure:example_grasp_generation}) which works like a ``magic wand'' to separate an object from its background using four classes of pixels: from the objects, probably from the object, not from the object, and probably not from the object. We determine the class of each pixel using three methods: HSV-based color segmentation, saliency based Segmentation \cite{DBLP:conf/cvpr/HouZ07}, Gaussian Mixture-based Background/Foreground Segmentation \cite{KaewTrakulPong2002AnIA}. When the three methods disagree, the class is set to ``probably object'' or ``probably background'' according to a few simple rules (Supplementary Sec. \ref{sec:si-grasp-generation}). We run a Canny edge detector \cite{DBLP:journals/pami/Canny86a} to obtain edges and we compute the skeleton \cite{DBLP:journals/cvgip/LeeKC94}.

 We first extract the object with the GrabCut algorithm  \cite{DBLP:journals/tog/RotherKB04} which works like a ``magic wand'' to separate an object from its background using four classes of pixels: from the objects, probably from the object, not from the object, and probably not from the object. We determine the class of each pixel using three methods: HSV-based color segmentation, saliency based Segmentation \cite{DBLP:conf/cvpr/HouZ07}, Gaussian Mixture-based Background/Foreground Segmentation \cite{KaewTrakulPong2002AnIA}. We run a Canny edge detector \cite{DBLP:journals/pami/Canny86a} to obtain edges and we compute the skeleton \cite{DBLP:journals/cvgip/LeeKC94}.

We generate grasps candidates by computing lines that are perpendicular to each edge pixel and to each skeleton pixel \cite{DBLP:journals/corr/abs-2001-02076}. This is achieved by looking at the two nearest neighbors of each point of the edges/skeleton. To keep the number of grasp candidates low, we skip edge/skeleton points if the distance with the previous point is below 4 pixels (Cornell dataset) and 8 pixels (our dataset). Finally, we add a random angle ($\pm 0.4$ rad, Gaussian distribution) to each candidate to increase the grasp diversity. At that stage, grasps have no gripper width since they are used to (1) generate patches (Section \ref{sec:grasp_representation}) which have a fixed size (therefore only the position and orientation matters), and (2) to query a GP, which will predict the gripper width using the latent representation of the patch.

% % In a similar fashion to \cite{DBLP:journals/corr/abs-2001-02076}, for each points of those two set of points, we compute their k closest neighbour. The points form a line, using linear regression we obtain an approximation of its slope. We find the slope of the line which is perpendicular see Fig. \ref{figure:grasp_representation} to this one by taking the opposite reciprocal slope, and convert it to the angle of the grasp $\theta = \arctan{(-1/slope)}$.

% Additionally we can add all the points part of the segmentation mask to the grasp candidates, setting their angle to their closest neighbour being a point part of the skeleton or the edges. To keep the number of grasp candidate reasonable, a minimum euclidean distance between each candidate is specified. And finally we add a random angle tweak to each candidate to improve grasp diversity. 

These $n$ points and angles $[x,y,\theta]^n$ are the list of grasps candidates for a specific RGB-D image. 

% \yoann{todo, add how much time it takes on average to generate graps that way on a image}

\subsection{Grasp candidate representation (b)}
\label{sec:grasp_representation}

% We encode grasps as 7-channel \emph{image patches} (Fig. \ref{figure:grasp_representation}), which were previously used in \cite{DBLP:conf/icra/PintoG16, DBLP:conf/rss/MahlerLNLDLOG17}, because they combine a specification of the grasp (width of the gripper, orientation, position relative to the object) with an image of the object to recognize it. These image patches are easy to feed to convolutional neural networks, by contrast to a coordinate-based or feature-based (orientation, position, ...) representation, which would need to be associated to the right object.

% To create this patches from $[x,y,\theta]$ (section \ref{sec:grasp_generation}), we: (1) rotate the image to correspond to the orientation of the segment, (2) translate the image to be centered on the center of the segment, (3) crop the image to $128\times128$ pixels. The width of the gripper is ignored for the patch extraction stage. Our images patches are 7-channel: 3 channels for the RGB image, 1 channel for the depth, and 3 channels for the surface normal image, generated from the depth gray-scale image.

% \subsection{Grasp representation as an oriented cropped data patch}\label{sec:grasp_representation}

We encode grasps as 7-channel \emph{image patches} (Fig. \ref{figure:grasp_representation}), like in \cite{DBLP:conf/icra/PintoG16} and \cite{DBLP:conf/rss/MahlerLNLDLOG17}, which centers each image on its respective grasp. The main benefit of this representation is that it combines a specification of the grasp (orientation, position relative to the object) with an image of the object to recognize it. These image patches are easily fed to convolutional neural networks, by contrast to a coordinate-based or feature-based representation, which would need to be associated with the right object.

To create these patches from $[x,y,\theta]$ (section \ref{sec:grasp_generation}), we: (1) rotate the image to correspond to the orientation of the segment, (2) translate the image to be centered on the center of the segment, (3) crop the image to $128\times128$ pixels. The width of the gripper is ignored for the patch extraction stage. Our image patches are 7-channel: 3 channels for the RGB image, 1 channel for the depth, and 3 channels for the surface normal image, generated from the depth gray-scale image.

\subsection{Latent space for grasp (c)}
\label{sec:vae}
We here assume that we have access to a large dataset of patches (called $d$) that is not labelled. We train a convolutional $\beta$-Variational Auto-Encoder \cite{DBLP:conf/iclr/HigginsMPBGBML17,DBLP:journals/corr/abs-1804-03599} using a large number of patches (at least $40,000$, depending on the dataset) generated from RGB-D images.
The last layer of the decoder uses a $\tanh$ activation function because we normalize our input to [-1, 1]. Given the trained VAE, the extraction of grasp patches and their encoding into the latent space takes less than 2 seconds per scene on average on our computer.

\subsection{Expert preference model learning (d) with Gaussian processes}
\label{sec:GP}
% method that learn fast, method able to learn differents grasps for user purposes, method that can be added to existing pipelne, exploit cheap unsupervised data
We now assume that we have access to a second dataset, for which a small set of patches have been labeled either as positive (selected by the expert) or negative (grasp to avoid). For the positive examples, the dataset also contains the gripper opening width selected by the expert.

We train a Gaussian processes classifier \cite{DBLP:books/lib/RasmussenW06} that takes as input the latent code for a patch, that is, a grasp candidate, and outputs a score $s$ between 0 and 1 that describes the probability that the grasp would be chosen by the expert. Using the positive example, we also train Gaussian process regressor \cite{DBLP:books/lib/RasmussenW06} that takes the same input and outputs a probability distribution of the width of the gripper. Compared to neural networks, Gaussian processes classifiers are more accurate when there are few data  \cite{DBLP:books/lib/RasmussenW06}, at the expense of a longer query/training time (query is $O(n^2)$ with $n$ the number of samples). In our tests, we observed a query time of less than 0.06 second per scene.

%with $m$ the latent space dimension set to 32
For each generated grasp (Sec.~\ref{sec:grasp_generation}), we first generate the corresponding patch, encode it to the latent space using the $\beta$-VAE decoder, then query the GP to obtain its score. We select the grasp candidate with the highest score. To scale the GP classifier to many samples without compromising on query time, we use ``Scalable Variational Gaussian Process classifiers'' \cite{DBLP:conf/aistats/HensmanMG15}, which is based on a variational inducing point framework, although any GP classifier could be used instead (for less than about 500 labels, a standard GP classifier works well). The gripper opening width $g_w$ is selected by querying the GP regressor with the selected patch (a standard GP is used for this as we only need a single query, compared to many queries for the classifier).

%

% Gaussian processes \cite{DBLP:books/lib/RasmussenW06} perform well on small data-sets but are not easily used with images. Since we have access to a large image data set, we can solve that issue by using deep learning method to learn a feature extraction function that can compress our images into an input vector suitable to a Gaussian process. Similarly to \cite{kingma2014semisupervised}, we use the encoder part of this unsupervised representation learning method as the feature extractor. On top of the encoder we train a Gaussian process classifier to assign a score to each grasps candidates. Using generative models as pretraining for classification tasks with self-supervised objectives has shown effective results \cite{chen2020generative}.

%We then select the grasp candidate with the highest score, using the same latent representation as input to a Gaussian Process regressor trained with the labeled gripper length from the Expert preferences dataset (using only the positive examples), giving us an estimation of the opening of the gripper. 

To use the oriented rectangle representation (see Fig. \ref{figure:grasp_representation}), we use a fixed \emph{height} value. When executing the selected grasp with the robot, the depth is computed by using the depth data from the RGB-D camera: we extract an oriented cropped patch of the depth point cloud with the width of the selected grasp and a fixed height (5 pixels), and we use the closest point to the gripper $g_d$ (that is, the highest point of the object) as the $z$-reference.

\section{Experimental evaluation}\label{sec:experiments}

\begin{figure*}
   \centering
   \includegraphics[width=\textwidth]{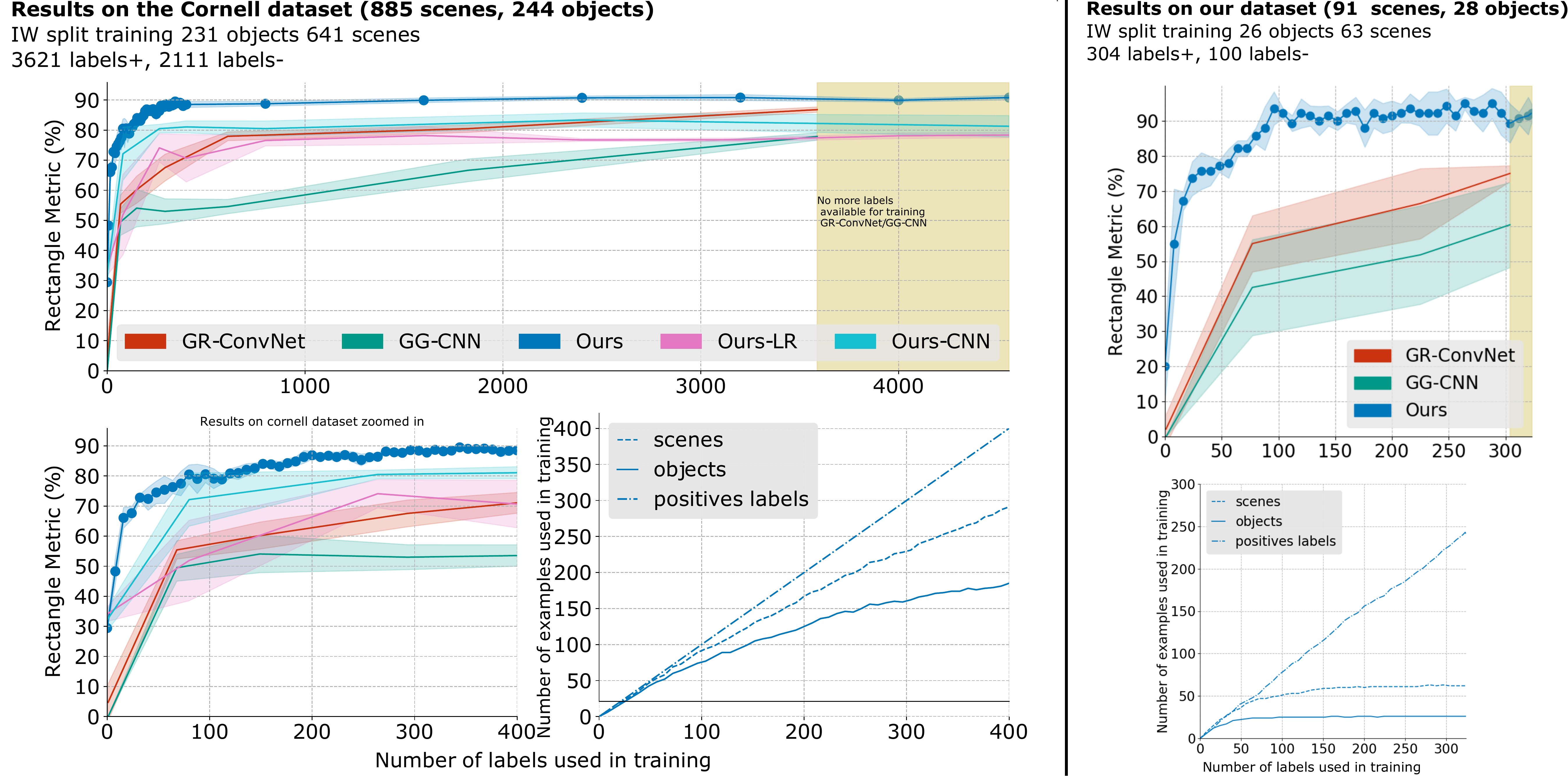}
   
   \caption{Rectangle metric for the Cornell dataset and our dataset.
   The $x$-axis corresponds to the number of labels used for training (training+validation data, $641$ scenes) and the $y$-axis to the rectangle metric in percent of the number of scenes in the test set ($244$ scenes). The rectangle metric counts how often the selected grasp matches at least one of the positive labels. For the Cornell dataset, the top plot shows the rectangle metric from $0$ to all the available labels (GR-ConvNet and GG-CNN do not use the negative labels, therefore they cannot use all the labels), the bottom left plot corresponds to the same data but focuses on the results from $0$ to $400$ labels, and the bottom right plot reports the number of scenes and objects with regards to the number of labels used for training. To better understand the performance of our pipeline, we compare two ``ablations'': switching the GP classifier with a logistic regression model for binary classification (``Ours-LR'') and using a CNN binary classifier instead of the VAE+GP combination (``Ours-CNN'').}
    \label{fig:Cornell_results}

\end{figure*}

We evaluate our approach on the Cornell dataset \cite{DBLP:conf/rss/LenzLS13} (885 scenes, 244 objects, 5055 positive label, 2822 negative labels) and on a custom dataset for which humans typically have preferences (91 scenes, 28 objects, 447 positive labels, 145 negative labels). For the Cornell dataset, the VAE 
%(Supplementary Table \ref{tab:dataset-vae}) 
is trained on the 885 scenes (244 objects). For our dataset, images are obtained with an Intel RealSense D415 Depth Camera mounted on the gripper of a Franka-Emika Panda robot (Fig. \ref{fig:hammer}) that is positioned 65 centimeters above the objects. The objects are mostly from the YCB dataset \cite{DBLP:journals/corr/CalliWSSAD15}. We used a 32-dimensional latent space (output of the VAE and input of the GP).

We focus on the performance with very few labels (fewer than 4000, ideally fewer than 50) because we envision interactive or semi-interactive scenarios in which an expert does not want to spend much time in labeling. In addition, we are interested in generalizing to new views of objects for which we have labels, and not necessarily to new objects that have never been seen. As a consequence, we make no effort to split the training set and the learning set into disjoint sets of objects; on the contrary, we typically expect our algorithm to select the right grasp with 1-2 grasp examples of the same object from different views. For instance, we do not expect our algorithm to know how to grasp a hammer if it has never seen a hammer, but we want it to learn how to grasp a hammer from any point of view once it has been explained how to do so once or twice.

%885 scenes (244 objects) are used for the Cornell dataset and 782 (124 objects) for the experiments with our datasets.

% and one in which objects are on a heap (41 scenes, 28 objects, 2504 positive labels, 1035 negative labels)

% \todo[inline]{
% What is the aim of the experiments? Start with a paragraph explaining what is the purpose of the experiments and what you want to show. Do you want to demonstrate that it works? That it works better than previous methods? Better in some conditions and worse in other conditions? Clarify what is the objective of the experiments.}

% \subsection{Comparison on the Cornell grasping dataset and to other grasping algorithm Dex-Net, GGCNN and GR-Convnet}
% Kumra etal.[Kumraet al., 2019]proposed a novel Generative Resid-ual Convolutional Neural Network (GR-ConvNet) model thatcan generate robust antipodal grasps from a n-channel imageinput.

We compare our method to two state-of-the-art grasping methods that are based on expert labels: Generative Grasping CNN (GG-CNN) \cite{DBLP:conf/rss/MorrisonLC18} and Generative Residual Convolutional Neural Network (GR-ConvNet) \cite{DBLP:journals/corr/abs-1909-04810} using the rectangle metric (Sec.~\ref{sec:proble_statement}). We were unable to run Dex-Net 4.0 \cite{Mahlereaau4984} on the Cornell dataset because the objects are too far from the camera (Dex-Net requires 0.5 to 0.7m) and the dataset does not give the camera intrinsic parameters. GG-CNN and GR-Convnet are two recent algorithms that use a fully-convolutional neural network to generate grasp quality and grasp poses/width ($x,y,\theta, w$) at every pixel from RGB-D scenes and to sample grasp candidates. GR-ConvNet reports a state-of-the-art accuracy of 97.7\% on the Cornell dataset. %To evaluate the contribution of the GP, we also compare to a baseline in which a logistic regression (from Scikit-Learn) is used in place of the GP.

For the following experiments, we are interested in comparing our method's performances (in terms of following the preferences) according to the number of available labels. 
To split our dataset image-wise, we order the RGB-D scenes by objects and randomly pick one scene per object for the testing data set, on which every algorithm is tested.

We generate the training and validation sets for a specified number of labeled grasps by randomly selecting data points outside of the test set. These labels are then split into 5 groups in order to perform 5-fold cross validation to train  GG-CNN and GR-ConvNet. Each method is trained with the same labels available for training, with this difference: GG-CNN and GR-ConvNet discard negative grasps for training (in these methods, everything that is not positive is considered as negative), and our method discards the validation data (since we do not need it for the GP). For instance, for 100 randomly selected labels (e.g., 60 positives and 40 negatives), each fold contains 80 labels for training and 20 for validation; our method is trained with 80 labels whereas GG-CNN and GR-ConvNet would use on average 48 labels (36 labels for training and 12 labels for validation). Please note that we report data using the number of used labels (training and validation), not the number of randomly selected labels, so that the comparison is as fair as possible.

%% TODO : what is a "data point"? 

% To illustrate a case where we would want to compare the method with 100 randomly selected labels, with for example 60 of them being positive and 40 negatives. Each fold would contain 80 labels for training and 20 for validation. Our method would be train with 80 labels, GG-CNN and GR-ConvNet would use on average 48 labels, 36 labels for training and 12 labels validation.

% Training on 0/100 to 100/100 of the training dataset  

% % \begin{tabular}{|*{5}{c|}}
% %     \hline
% %      $D$ Heap  &  scenes & objects & + labels & - labels\\
% %     \hline
% %      Total:  & 91 & 28 & 211 & 243\\
% %     \hline
% %      Training:  & 63 & 26 & 151 & 158\\
% %     \hline
% %      Testing:  & 28 & 28 & 60 & 85\\
% %     \hline
% % \end{tabular}

% \begin{tabular}{*{5}{c}} \toprule
%     $D$ Heap  &  scenes & objects & + labels & - labels\\ \midrule
%     Total:  & 91 & 28 & 211 & 243\\
%     Training:  & 63 & 26 & 151 & 158\\
%      Testing:  & 28 & 28 & 60 & 85\\
% \end{tabular}

We run our method 5 times for each number of labels (each run is independent); we test with $0,10,20,\cdots,500$ labels and $1000, 2000, \cdots, 4558$ labels. Overall, we therefore launch $280$ ($56 \times 5 = 280$) independent runs of our learning algorithm (and therefore $280$ tests). For GR-ConvNet and GG-CNN, we test with $0, 100, 300, 400, 1000, 3000$, and $3588$ labels (these methods only use the positive labels) and 5 replicates (total $45$ runs for each baseline).  We train the VAE part of our method once for each dataset $\mathcal{D}$, using a 70\%-30\% training and validation split. The VAE network is trained for 48 hours on a Nvidia GTX 1080 Ti using the Keras framework and the Adam optimizer with the following parameters:  $|B|=64, \eta =0.001, w=128,h=128,\beta=1.0,m=32$. GG-CNN and GR-ConvNet are both trained on Nvidia GTX 1080 with the provided settings and data augmentations suited for the Cornell Dataset.

\subsection{Qualitative results}
\label{sec:qualitative}
We first checked that our method fits our expectations on an object with a clear choice (Fig. \ref{fig:hammer}). In this experiment, a (toy) hammer has to be taken from the top of the handle (close to the head), and not from the bottom (far from the head). We have chosen this example because a hammer has a very non-uniform distribution of mass that cannot be deduced from the shape alone: grasps that are not close to the head are unlikely to be successful.

For this preliminary experiment, we used a single image with two labels: a positive label for the top part of the handle and a negative label for the bottom part (Fig. \ref{fig:hammer}). We then captured 21 additional scenes of the same hammer in different positions and evaluated how often our method chose to grasp the hammer from the top part  (to perform this evaluation, we labeled the 21 scenes, but did not use these labels in training).

The results show that our method selects the right grasp in 19/21 scenes 
%(Supplementary Fig.~\ref{fig:hammer-sup}) 
and only fails in the 2 scenes in which the hammer is upright on the table (for which the handle is not accessible with a top-down grasp).

\subsection{Quantitative results}
\label{sec:cornell_results}

After about 16 example scenes from the Cornell dataset (16 random labels, which correspond to about 13 positive examples for about 14 objects), our method selects the right grasp (according to the rectangle metric) for 66\% of the test scenes (244 scenes) (Fig.~\ref{fig:Cornell_results}) that include many unseen objects (244 objects total, whereas at best 16 different objects have been seen). As a comparison point, randomly selecting the grasp among those generated leads to a score of 29\% (this corresponds to the ``0 label'' value on the plot), which means that the GP classifier is learning useful knowledge. After 68 labels, our method accurately predicts the grasps for 76\% of the scenes, whereas GR-ConvNet selects the right grasp in only $55\%$ of the test scenes, and GG-ConvNet only $49$\%. Our method exceeds 80\% with 80 labels (63 different objects, 71 scenes, 54 positive labels) and 89.5\% with 344 labels. By contrast, the best score reached by GR-ConvNet after 3588 labels\footnote{Please note that the authors GR-ConvNet \cite{DBLP:journals/corr/abs-1909-04810} report an accuracy of $97.7$\% with all the available labels, but we were unable to obtain the same score with their code (they do not report any score with a subset of the Cornell dataset, as we do here). Our results are consistent with the results reported in the GG-CNN literature \cite{DBLP:journals/corr/abs-1909-04810}.} is $86.8$\% and by GG-CNN is $77.9$\%.  Overall, our method clearly outperforms the best score of all the baselines after only 344 labels (versus more than $3000$ for the baselines), and only a few hundred labels are enough to get more than 80\% of the grasps right.

\begin{figure}[t]
   \centering
   \includegraphics[width=\linewidth]{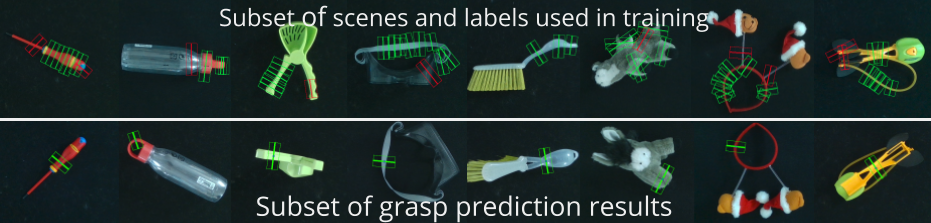}
    \caption{Typical results from our dataset: 
    the top row shows the objects in scenes used for training (additional scenes of these objects were also used), and the bottom row shows some of our predicted grasps on unseen scenes of the same object. For benchmarking data-efficiency, a random subset of labels is used (e.g., only 2 labels for each object). 
    %The predictions follows the arbitrary rules from the expert for each objects: some part of the object is allowed (the handle of the screwdriver, the red bottle cap, the glasses string, ...) and some other is not (the brush hairs, the head of the donkey plush, the fin of the torpedo, ...). 
    %More details are available in appendix \ref{sec:datasets}
    }
    \label{figure:Heap_predicted_grasps}
\end{figure}

The results are similar for our dataset (Fig.~\ref{figure:Heap_predicted_grasps}), which is smaller but similar to the Cornell dataset (single object with labels)
%, Supplementary table \ref{tab:dataset-labels})
: our method reaches a score of $74$\% with only 24 labels (from 22 scenes, 15 different objects, 20 of them being positive), and 93.57\% with 96 labels. To provide a reference point, with 300 labels, GR-ConvNet obtains a score of $75$\% and GG-CNN a score of $60$\%.

\subsection{Ablation study}
Our pipeline combines a VAE to learn a generic latent space with a GP to classify grasp candidates in this latent space. To understand the contribution of the GP, we replaced it with a logistic regression classifier (from Scikit-Learn). The results (Fig. \ref{fig:Cornell_results}, ``Ours-LR'') show that the logistic regression leads to scores similar to those obtained by GR-ConvNet, which shows that the GP classifier is a key component. To evaluate the contribution of the VAE trained on many unlabeled images, we then trained a CNN classifier with the same architecture as the VAE (it takes the same patches as input) but that outputs directly the class of the grasp candidate (correct or incorrect). This CNN is trained only on the labeled patches. The results show  (Fig. \ref{fig:Cornell_results}, ``Ours-CNN'') that this simple approach can lead to competitive results but is clearly outperformed by our pipeline. Overall, both training on unlabeled images and using a GP are necessary to reach a good score according to the rectangle metric.

\subsection{Tests with a Franka-Emika Panda robot}
The supplementary video shows a Franka-Emika Panda robot that performs the grasps learned with our pipeline and our dataset (Fig.~\ref{fig:hammer}). For each of the 28 objects, the camera on the gripper is positioned 65 cm above the table, our pipeline selects the grasp, and the robot executes it using the planning algorithms of the MoveIt library \cite{chitta2012moveit}. 

The objective of our method is to follow the expert preferences, not to achieve the most successful grasps: the expert might require a grasp that is much harder than other grasps for their own reason (which is what the rectangle metrics evaluate). In these robot experiments, 17\% of the grasps did not match what the expert expected and 20\% of the grasps were unsuccessful (these two sets of failure do not fully overlap). Most failures came from issues with the grasp generation step (e.g., the preferred grasp was not proposed by the generator or no good grasp was proposed).

\section{CONCLUSIONS}
Our pipeline is highly data-efficient because it uses a shape-based grasp generator as a prior, instead of learning grasps only from labels. A direct consequence is that our pipeline is at least as good as the grasp generator that is used: it always chooses among the proposed grasps. When more effective grasp generators are developed, they can be directly leveraged to both create the latent space and the candidates, and our pipeline becomes more effective.

This paper focuses on offline training in order to compare it to the state-of-the-art on well-defined datasets. However, our pipeline would fit well in online scenarios during which a ``supervisor'' corrects the robot only whenever it is wrong, that is, when the shape-based grasp generator is wrong. The supervisor would then act like a teacher with a competent student, with minimal supervision. In future work, we will design human-in-the-loop studies in conjunction with recent grasp generators to assess the performance on such an online learning process.

% \todo[inline]{The conclusion should summarize the main points of the paper - BUT DO NOT COPY THE ABSTRACT! What was the main contribution of our work? Why was the work important? Even if it is a first step in the big picture of the problem that must be solved, why is this step important? {\bf What is the take home message?} What are the next steps and things we still need to solve? Maybe we made assumptions that we do not want for the next steps? How will we do it? Sketch the idea of the next steps.}

%\clearpage
% % The acknowledgments are automatically included only in the final version of the paper.
% \acknowledgments{}

%===============================================================================

% Note: RA Letters do not include biographical sketches nor photographs of authors. Figures, Tables, and References In all submissions, all figures and tables must be numbered and cited in the text. References must be in a separate reference section at the end of the paper, with items referred to by numerals in square brackets. References must be completed in IEEE style as follows:

%     Author(s), first initials followed by last name, title in quotation marks, periodical, volume, inclusive page numbers, month and year.
%     Books: Author(s), first initials followed by last name, title, location, publisher, year, chapter, page numbers.

%\printbibliography
\bibliographystyle{IEEEtran}
\bibliography{references.bib}

\begin{thebibliography}{10}
\def\url{}
\csname url@samestyle\endcsname
\providecommand{\newblock}{\relax}
\providecommand{\bibinfo}[2]{#2}
\providecommand{\BIBentrySTDinterwordspacing}{\spaceskip=0pt\relax}
\providecommand{\BIBentryALTinterwordstretchfactor}{4}
\providecommand{\BIBentryALTinterwordspacing}{\spaceskip=\fontdimen2\font plus
\BIBentryALTinterwordstretchfactor\fontdimen3\font minus
  \fontdimen4\font\relax}
\providecommand{\BIBforeignlanguage}[2]{{%
\expandafter\ifx\csname l@#1\endcsname\relax
\typeout{** WARNING: IEEEtran.bst: No hyphenation pattern has been}%
\typeout{** loaded for the language `#1'. Using the pattern for}%
\typeout{** the default language instead.}%
\else
\language=\csname l@#1\endcsname
\fi
#2}}
\providecommand{\BIBdecl}{\relax}
\BIBdecl

\bibitem{prattichizzo2016grasping}
D.~Prattichizzo and J.~C. Trinkle, ``Grasping,'' in \emph{Springer handbook of
  robotics}.\hskip 1em plus 0.5em minus 0.4em\relax Springer, 2016, pp.
  955--988.

\bibitem{du2021vision}
G.~Du, K.~Wang, S.~Lian, and K.~Zhao, ``Vision-based robotic grasping from
  object localization, object pose estimation to grasp estimation for parallel
  grippers: a review,'' \emph{Artificial Intelligence Review}, vol.~54, no.~3,
  pp. 1677--1734, 2021.

\bibitem{Mahlereaau4984}
J.~Mahler, M.~Matl, V.~Satish, M.~Danielczuk, B.~DeRose, S.~McKinley, and
  K.~Goldberg, ``Learning ambidextrous robot grasping policies,'' \emph{Science
  Robotics}, vol.~4, no.~26, 2019.

\bibitem{DBLP:conf/rss/MahlerLNLDLOG17}
J.~Mahler, J.~Liang, S.~Niyaz, M.~Laskey, R.~Doan, X.~Liu, J.~A. Ojea, and
  K.~Goldberg, ``Dex-net 2.0: Deep learning to plan robust grasps with
  synthetic point clouds and analytic grasp metrics,'' in \emph{Robotics:
  Science and Systems XIII}, 2017.

\bibitem{vae_kingma_2014_iclr}
D.~P. {Kingma} and M.~{Welling}, ``Auto-encoding variational bayes,'' in
  \emph{International Conference on Learning Representations (ICLR)}, 2014.

\bibitem{DBLP:journals/corr/abs-1804-03599}
\BIBentryALTinterwordspacing
C.~P. Burgess, I.~Higgins, A.~Pal, L.~Matthey, N.~Watters, G.~Desjardins, and
  A.~Lerchner, ``Understanding disentangling in {\(\beta\)}-vae,'' \emph{CoRR},
  vol. abs/1804.03599, 2018.  \url{http://arxiv.org/abs/1804.03599}
\BIBentrySTDinterwordspacing

\bibitem{DBLP:conf/aistats/HensmanMG15}
J.~Hensman, A.~G. de~G.~Matthews, and Z.~Ghahramani, ``Scalable variational
  gaussian process classification,'' in \emph{Proceedings of the Eighteenth
  International Conference on Artificial Intelligence and Statistics,
  {AISTATS}}, vol.~38, 2015.

\bibitem{DBLP:books/lib/RasmussenW06}
C.~E. Rasmussen and C.~K.~I. Williams, \emph{Gaussian processes for machine
  learning}.\hskip 1em plus 0.5em minus 0.4em\relax {MIT} Press, 2006.

\bibitem{quigley2009ros}
M.~Quigley, K.~Conley, B.~Gerkey, J.~Faust, T.~Foote, J.~Leibs, R.~Wheeler,
  A.~Y. Ng \emph{et~al.}, ``{ROS}: an open-source robot operating system,'' in
  \emph{ICRA workshop on open source software}, 2009.

\bibitem{yoo2017variational}
Y.~Yoo, S.~Yun, H.~Jin~Chang, Y.~Demiris, and J.~Young~Choi, ``Variational
  autoencoded regression: high dimensional regression of visual data on complex
  manifold,'' in \emph{Proceedings of the IEEE Conference on Computer Vision
  and Pattern Recognition (CVPR)}, 2017.

\bibitem{DBLP:conf/icra/JiangMS11}
Y.~Jiang, S.~Moseson, and A.~Saxena, ``Efficient grasping from {RGBD} images:
  Learning using a new rectangle representation,'' in \emph{{IEEE}
  International Conference on Robotics and Automation {(ICRA)}}, 2011.

\bibitem{DBLP:conf/icra/FerrariC92}
C.~Ferrari and J.~F. Canny, ``Planning optimal grasps,'' in \emph{{IEEE}
  International Conference on Robotics and Automation ({ICRA)}}, 1992, pp.
  2290--2295.

\bibitem{DatadrivenGrasp2013Kragic}
J.~Bohg, A.~Morales, T.~Asfour, and D.~Kragic, ``Data-driven grasp synthesis
  --- a survey,'' \emph{IEEE Transactions on Robotics}, vol.~30, no.~2, pp.
  289--309, 2013.

\bibitem{DBLP:conf/iros/JohnsLD16}
E.~Johns, S.~Leutenegger, and A.~J. Davison, ``Deep learning a grasp function
  for grasping under gripper pose uncertainty,'' in \emph{{IEEE/RSJ}
  International Conference on Intelligent Robots and Systems, (IROS)}, 2016.

\bibitem{DBLP:conf/corl/GualtieriP18}
M.~Gualtieri and R.~P. Jr., ``Learning 6-dof grasping and pick-place using
  attention focus,'' in \emph{2nd Annual Conference on Robot Learning (CoRL)},
  vol.~87.\hskip 1em plus 0.5em minus 0.4em\relax {PMLR}, 2018, pp. 477--486.

\bibitem{DBLP:conf/iros/ZengSWLRF18}
A.~Zeng, S.~Song, S.~Welker, J.~Lee, A.~Rodriguez, and T.~A. Funkhouser,
  ``Learning synergies between pushing and grasping with self-supervised deep
  reinforcement learning,'' in \emph{{IEEE/RSJ} International Conference on
  Intelligent Robots and Systems ({IROS})}, 2018.

\bibitem{DBLP:conf/iccv/MousavianEF19}
A.~Mousavian, C.~Eppner, and D.~Fox, ``{6-DOF GraspNet}: Variational grasp
  generation for object manipulation,'' in \emph{{IEEE/CVF} International
  Conference on Computer Vision ({ICCV})}, 2019, pp. 2901--2910.

\bibitem{DBLP:conf/corl/ViereckPSP17}
U.~Viereck, A.~ten Pas, K.~Saenko, and R.~P. Jr., ``Learning a visuomotor
  controller for real world robotic grasping using simulated depth images,'' in
  \emph{Conference on Robot Learning (CoRL)}, 2017.

\bibitem{DBLP:conf/icra/BousmalisIWBKKD18}
K.~Bousmalis, A.~Irpan, P.~Wohlhart, Y.~Bai, M.~Kelcey, M.~Kalakrishnan,
  L.~Downs, J.~Ibarz, P.~Pastor, K.~Konolige, S.~Levine, and V.~Vanhoucke,
  ``Using simulation and domain adaptation to improve efficiency of deep
  robotic grasping,'' in \emph{{IEEE} International Conference on Robotics and
  Automation {(ICRA)}}, 2018, pp. 4243--4250.

\bibitem{DBLP:conf/icra/PintoG16}
L.~Pinto and A.~Gupta, ``Supersizing self-supervision: Learning to grasp from
  50k tries and 700 robot hours,'' in \emph{{IEEE} International Conference on
  Robotics and Automation {(ICRA)}}, 2016.

\bibitem{DBLP:journals/ijrr/LevinePKIQ18}
S.~Levine, P.~Pastor, A.~Krizhevsky, J.~Ibarz, and D.~Quillen, ``Learning
  hand-eye coordination for robotic grasping with deep learning and large-scale
  data collection,'' \emph{International Journal of Robotics Research},
  vol.~37, no. 4-5, pp. 421--436, 2018.

\bibitem{DBLP:journals/arobots/DanielKVM015}
C.~Daniel, O.~Kroemer, M.~Viering, J.~Metz, and J.~Peters, ``Active reward
  learning with a novel acquisition function,'' \emph{Autonomous Robots},
  vol.~39, no.~3, pp. 389--405, 2015.

\bibitem{DBLP:conf/icra/PinslerAO0N18}
R.~Pinsler, R.~Akrour, T.~Osa, J.~Peters, and G.~Neumann, ``Sample and feedback
  efficient hierarchical reinforcement learning from human preferences,'' in
  \emph{{IEEE} International Conference on Robotics and Automation, {(ICRA)}},
  2018, pp. 596--601.

\bibitem{DBLP:journals/scirobotics/MigliozziLFS19}
A.~Migliozzi, G.~Laudante, P.~Falco, and B.~Siciliano, ``Vision-based grasp
  learning of an anthropomorphic hand-arm system in a synergy-based control
  framework,'' \emph{Science Robotics}, vol.~4, no.~26, 2019.

\bibitem{DBLP:conf/hri/PraveenaSMG19}
P.~Praveena, G.~Subramani, B.~Mutlu, and M.~Gleicher, ``Characterizing input
  methods for human-to-robot demonstrations,'' in \emph{14th {ACM/IEEE}
  International Conference on Human-Robot Interaction {(HRI)}}, 2019.

\bibitem{DBLP:journals/corr/abs-1912-04344}
\BIBentryALTinterwordspacing
S.~Song, A.~Zeng, J.~Lee, and T.~A. Funkhouser, ``Grasping in the wild:
  Learning {6DoF} closed-loop grasping from low-cost demonstrations,''
  \emph{CoRR}, vol. abs/1912.04344, 2019.
  \url{http://arxiv.org/abs/1912.04344}
\BIBentrySTDinterwordspacing

\bibitem{chu2018deep}
F.~Chu, R.~Xu, and P.~A. Vela, ``Real-world multiobject, multigrasp
  detection,'' \emph{IEEE Robotics and Automation Letters}, 2018.

\bibitem{DBLP:conf/rss/MorrisonLC18}
D.~Morrison, J.~Leitner, and P.~Corke, ``Closing the loop for robotic grasping:
  {A} real-time, generative grasp synthesis approach,'' in \emph{Robotics:
  Science and Systems XIV}, 2018.

\bibitem{DBLP:journals/corr/abs-1909-04810}
\BIBentryALTinterwordspacing
S.~Kumra, S.~Joshi, and F.~Sahin, ``Antipodal robotic grasping using generative
  residual convolutional neural network,'' \emph{CoRR}, vol. abs/1909.04810,
  2019.  \url{http://arxiv.org/abs/1909.04810}
\BIBentrySTDinterwordspacing

\bibitem{DBLP:conf/icra/ZengSYDHBMTLRFA18}
A.~Zeng, S.~Song \emph{et~al.}, ``Robotic pick-and-place of novel objects in
  clutter with multi-affordance grasping and cross-domain image matching,'' in
  \emph{{IEEE} International Conference on Robotics and Automation {(ICRA)}},
  2018.

\bibitem{de2020learning}
E.~De~Coninck, T.~Verbelen, P.~Van~Molle, P.~Simoens, and B.~Dhoedt, ``Learning
  robots to grasp by demonstration,'' \emph{Robotics and Autonomous Systems},
  vol. 127, p. 103474, 2020.

\bibitem{lenz2015deep}
I.~Lenz, H.~Lee, and A.~Saxena, ``Deep learning for detecting robotic grasps,''
  \emph{The International Journal of Robotics Research}, vol.~34, no. 4-5, pp.
  705--724, 2015.

\bibitem{park2018classification}
D.~Park and S.~Y. Chun, ``Classification based grasp detection using spatial
  transformer network,'' \emph{arXiv preprint arXiv:1803.01356}, 2018.

\bibitem{van2018learning}
P.~Van~Molle, T.~Verbelen, E.~De~Coninck, C.~De~Boom, P.~Simoens, and
  B.~Dhoedt, ``Learning to grasp from a single demonstration,'' \emph{arXiv
  preprint arXiv:1806.03486}, 2018.

\bibitem{DBLP:conf/ro-man/HelenonBNTG20}
F.~H{\'{e}}l{\'{e}}non, L.~Bimont, E.~Nyiri, S.~Thiery, and O.~Gibaru,
  ``Learning prohibited and authorised grasping locations from a few
  demonstrations,'' in \emph{29th {IEEE} International Conference on Robot and
  Human Interactive Communication {(RO-MAN)}}, 2020, pp. 1094--1100.

\bibitem{DBLP:conf/iros/DepierreD018}
A.~Depierre, E.~Dellandr{\'{e}}a, and L.~Chen, ``Jacquard: {A} large scale
  dataset for robotic grasp detection,'' in \emph{2018 {IEEE/RSJ} International
  Conference on Intelligent Robots and Systems {(IROS)}}, 2018.

\bibitem{DBLP:journals/corr/abs-2001-02076}
\BIBentryALTinterwordspacing
M.~Vohra, R.~Prakash, and L.~Behera, ``Real-time grasp pose estimation for
  novel objects in densely cluttered environment,'' \emph{CoRR}, vol.
  abs/2001.02076, 2020.  \url{http://arxiv.org/abs/2001.02076}
\BIBentrySTDinterwordspacing

\bibitem{DBLP:conf/icra/MahlerPHRLAKKKG16}
J.~Mahler, F.~T. Pokorny, B.~Hou, M.~Roderick, M.~Laskey, M.~Aubry,
  K.~Kohlhoff, T.~Kr{\"{o}}ger, J.~J. Kuffner, and K.~Goldberg, ``Dex-net 1.0:
  {A} cloud-based network of 3d objects for robust grasp planning using a
  multi-armed bandit model with correlated rewards,'' in \emph{{IEEE}
  International Conference on Robotics and Automation, {(ICRA)}}, 2016.

\bibitem{DBLP:journals/tog/RotherKB04}
C.~Rother, V.~Kolmogorov, and A.~Blake, ``"grabcut": interactive foreground
  extraction using iterated graph cuts,'' \emph{{ACM} Transactions on
  Graphics}, vol.~23, no.~3, pp. 309--314, 2004.

\bibitem{DBLP:conf/cvpr/HouZ07}
X.~Hou and L.~Zhang, ``Saliency detection: {A} spectral residual approach,'' in
  \emph{{IEEE} Computer Society Conference on Computer Vision and Pattern
  Recognition {(CVPR})}, 2007.

\bibitem{KaewTrakulPong2002AnIA}
P.~KaewTraKulPong and R.~Bowden, ``An improved adaptive background mixture
  model for real-time tracking with shadow detection,'' in \emph{Proc. 2nd
  European Workshop on Advanced Video Based Surveillance Systems}, 2002.

\bibitem{DBLP:journals/pami/Canny86a}
J.~F. Canny, ``A computational approach to edge detection,'' \emph{{IEEE}
  Transactions on Pattern Analysis and Machine Intelligence}, vol.~8, no.~6,
  pp. 679--698, 1986.

\bibitem{DBLP:journals/cvgip/LeeKC94}
T.~Lee, R.~L. Kashyap, and C.~Chu, ``Building skeleton models via 3-d medial
  surface/axis thinning algorithms,'' \emph{Graphical Models and Image
  Processing}, vol.~56, no.~6, pp. 462--478, 1994.

\bibitem{DBLP:conf/iclr/HigginsMPBGBML17}
I.~Higgins, L.~Matthey, A.~Pal, C.~Burgess, X.~Glorot, M.~Botvinick,
  S.~Mohamed, and A.~Lerchner, ``Beta-vae: Learning basic visual concepts with
  a constrained variational framework,'' in \emph{5th International Conference
  on Learning Representations {(ICLR)}}, 2017.

\bibitem{DBLP:conf/rss/LenzLS13}
I.~Lenz, H.~Lee, and A.~Saxena, ``Deep learning for detecting robotic grasps,''
  in \emph{Robotics: Science and Systems IX}, 2013.

\bibitem{DBLP:journals/corr/CalliWSSAD15}
\BIBentryALTinterwordspacing
B.~{\c{C}}alli, A.~Walsman, A.~Singh, S.~S. Srinivasa, P.~Abbeel, and A.~M.
  Dollar, ``Benchmarking in manipulation research: The {YCB} object and model
  set and benchmarking protocols,'' \emph{CoRR}, vol. abs/1502.03143, 2015.
  \url{http://arxiv.org/abs/1502.03143}
\BIBentrySTDinterwordspacing

\bibitem{chitta2012moveit}
S.~Chitta, I.~Sucan, and S.~Cousins, ``{MoveIt!}'' \emph{IEEE Robotics \&
  Automation Magazine}, vol.~19, no.~1, pp. 18--19, 2012.

\end{thebibliography}

\end{document}